\documentclass{article} 
\usepackage{iclr2020_conference,times}

\usepackage{amsmath,amsfonts,bm}

\def\eqref#1{equation~\ref{#1}}

\def\1{\bm{1}}

\DeclareMathAlphabet{\mathsfit}{\encodingdefault}{\sfdefault}{m}{sl}
\SetMathAlphabet{\mathsfit}{bold}{\encodingdefault}{\sfdefault}{bx}{n}

\usepackage{hyperref}
\usepackage{url}
\usepackage{graphicx}
\usepackage{booktabs}       

\title{Learning To Solve Differential Equations Across Initial Conditions}

\author{Shehryar Malik\thanks{Both authors contributed equally.} \\
Information Technology University\\
Lahore, Pakistan \\
\texttt{shehryar.malik@itu.edu.pk} \\
\And 
Usman Anwar\footnotemark[1] \\
Information Technology University\\
Lahore, Pakistan \\
\texttt{usman.anwar@itu.edu.pk} \\
\And
Ali Ahmed \\
Department of Electrical Engineering \\
Information Technology University\\
Lahore, Pakistan \\
\texttt{ali.ahmed@itu.edu.pk} \\
\And
Alireza Aghasi \\
J. Mack Robinson College of Business \\
Georgia State University \\
Atlanta, GA, USA. \\
\texttt{aaghasi@gsu.edu} \\
}

\iclrfinalcopy 

\begin{document}
\maketitle
\begin{abstract}
Recently, there has been a lot of interest in using neural networks for solving partial differential equations. A number of neural network-based partial differential equation solvers have been formulated which provide performances equivalent, and in some cases even superior, to classical solvers. However, these neural solvers, in general, need to be retrained each time the initial conditions or the domain of the partial differential equation changes. In this work, we posit the problem of approximating the solution of a fixed partial differential equation for any arbitrary initial conditions as learning a conditional probability distribution. We demonstrate the utility of our method on Burger's Equation.
\end{abstract}

\section{Introduction}
Partial differential equations (PDEs) are of great importance in various fields such as science, engineering and economics. However, despite this, it is generally not possible to obtain analytic solutions for them. Instead, one has to resort to numerical schemes for approximating these solutions. However, these numerical schemes are both slow and computationally intensive especially for higher dimensions. Furthermore, the higher the dimension is, the greater is the error in the calculation of the derivatives required to approximate the solution.

Recently, \cite{PINN} proposed to use neural networks to approximate solutions of PDEs. They do so by forcing neural networks to produce outputs that satisfy the PDE. The derivatives required to enforce this condition are computed using automatic differentiation and are exact up to the precision of the computing machine. Furthermore, this approach is \textit{mesh free} i.e. one does not need to discretize the domain of the solution as required in methods such as finite element analysis.

In parallel, we have seen tremendous improvements in the generalization capacities of machine learning methods such as generative adversarial networks and flow-based models. These models have demonstrated a remarkable ability to capture probability distributions even when only representative samples are available [\cite{glow}, \cite{biggan}, \cite{wavenet}].

In this work, we generalize the framework of \cite{PINN-GAN1} over the distribution of initial conditions and demonstrate that even if the model is trained on a subset of them, it is able to generalize well and produce solutions for any arbitrary initial conditions. Furthermore, our model also provides uncertainty quantifications which loosely correlate with the error in the solution.

This paper is structured as follows: in Section \ref{rw}, we provide an overview of related works on partial differential equations. Section \ref{method} presents our method and Section \ref{exp} talks about our experiments and results. Finally, Section \ref{conclusion} concludes the paper.

\section{Related Works}
\label{rw}
PDEs have been traditionally solved through numerical methods such as finite differences, finite element methods and spectral methods (see for e.g. \cite{classical-review}). \cite{1998} made the first attempt at using neural networks to approximate solutions of PDEs. However, the idea did not catch up. The recent renaissance of neural networks and its remarkable successes in computer vision and natural language processing [\cite{imagenet}, \cite{bert}] have sparked a new interest in scientific machine learning i.e. in applying data driven methods to problems in natural sciences [\cite{high-dimensional-pde}, \cite{protein-folding}, \cite{brain-atlas}]. Since a huge number of problems in the natural and physical sciences are described by PDEs, this has inevitably aroused interest in using machine learning to solve PDEs. In this regard, the physics-informed neural network method in \cite{PINN} was a critical development. It made two major contributions: (a) it showed that the automatic differentiation machinery of machine learning frameworks such as TensorFlow allowed the computation of partial derivatives to a higher order of accuracy than finite differences, and (b) it empirically showed that neural networks can be used to approximate solutions of PDEs by simply forcing them to produce outputs that satisfy the PDE. This framework has since been extended to fractional and stochastic PDEs [\cite{fPINN}, \cite{stochastic-pde}].

Neural network based solvers do not, in general, provide any convergence guarantees. Hence, it is important to have an uncertainty estimate for the solution produced by 
these methods. \cite{PINN-GAN1} and \cite{PINN-GAN2} formulate the method in \cite{PINN} in the framework of generative adversarial networks [\cite{GAN}] 
and show that it is possible to obtain uncertainty estimates as well.

Concurrent to this has been considerable work on combining various classical solvers with machine learning methods to obtain more efficient solvers such as combining the Runge Kutta method with convolutional networks in \cite{conv-RK} and the Galarkin method with deep learning in \cite{DGM}.

However, one area which has attracted considerably less attention is attempting to learn a `general' PDE solver through neural networks. General, in this context, refers to a PDE solver which does not need to be retrained if one or more constraints such as the initial or boundary condition or the domain are changed. Rather, the general PDE solver relies on the generalization capacity of neural networks to approximate the solution of a PDE across all possible sets of constraints. In this regard, \cite{neural-pde-solver} and \cite{gan-poisson} have, for example, demonstrated that it is possible to learn a general PDE solver for some simple linear and elliptic PDEs. However, developing a single general PDE solver for all types of PDEs (parabolic, elliptic and hyperbolic) still remains an open problem.

\section{Methodology}
\label{method}
We consider partial differential equations (PDEs) of the form 
\begin{equation}
    \frac{\partial u(\mathbf{x},t; i)}{\partial t} = f\left(\mathbf{x},t,u(\mathbf{x},t;i),\frac{\partial u(\mathbf{x},t;i)}{\partial \mathbf{x}},\ldots,\frac{\partial^d u(\mathbf{x},t;i)}{\partial \mathbf{x}^d}\right)
    \label{eqn:pde}
\end{equation}
where $\mathbf{x} \in \mathbb{R}^p$, $t \in \mathbb{R}$, $u: \mathbb{R}^p \times \mathbb{R} \rightarrow \mathbb{R}$ is the solution of the PDE and $f$ is a known function. Since the solution $u$ of this PDE depends on the initial conditions $i$, we make this dependence explicit by writing $u(\mathbf{x},t;i)$. We denote the distribution of initial conditions with $\mathcal{I}$.

We propose to train a single Generative Adversarial Network (GAN) [\cite{GAN}] for solving a PDE of the form of Equation \ref{eqn:pde} for any initial conditions drawn from the distribution $\mathcal{I}$. Note that the PDE is fixed (i.e. $f$ is fixed) and only the initial conditions are varied. The generator $G_\theta$ takes in as input a random noise vector $\mathbf{z} \sim \mathcal{N}(\mathbf{0},\mathbf{1})$, $\mathbf{x}$, $t$ and a particular instance of the initial conditions $i \sim \mathcal{I}$ and generates an approximation of $u(\mathbf{x},t;i)$. The discriminator $D_\psi$ takes two tuples $(\mathbf{x}_1, t_1, G_\theta(\mathbf{x}_1, t_1, i_1, \mathbf{z}_1),i_1)$ and $(\mathbf{x}_2, t_2, u(\mathbf{x}_2, t_2; i_2), i_2)$ and is asked to identify the tuple generated by the generator (as in a typical GAN setup).

The generator $G_\theta$ is composed of three networks: the encoder, approximator and reconstructor. The encoder takes in as input a particular instance of the initial conditions $i \sim \mathcal{I}$ and encodes it into some latent vector $v$. The approximator then takes in as input the latent vector $v$ along with some spatio-temporal coordinates $(\mathbf{x},t)$ and outputs an approximation of $u(\mathbf{x},t;i)$. Finally, the reconstructor takes in as input the outputs of the approximator for several different spatio-temporal coordinates but for the same initial conditions and tries to reconstruct the latent representation of the initial conditions $v$ from these approximations. The reconstructor (inspired, in part, by the InfoGAN architecture [\cite{info-gan}]) forces the approximator to condition \textit{all} of its outputs on the initial conditions, since otherwise the approximator may learn to produce samples from just one field irrespective of the initial conditions provided to it.

In addition to minimizing the typical GAN loss [\cite{GAN}] given by
\begin{equation}
    \mathcal{L}_{GAN} = \mathbb{E}_{i,\mathbf{x},t,\mathbf{z}}\left[\log D_\psi\left(\mathbf{x},t,u(\mathbf{x},t;i),i\right)\right] + \mathbb{E}_{i,\mathbf{x},t,\mathbf{z}}\left[1-\log\left(D_\psi\left(\mathbf{x},t,G_\theta(\mathbf{x},t,i,\mathbf{z}),i\right)\right)\right],
\end{equation}
we force the samples produced by the generator to satisfy Equation \ref{eqn:pde} by minimizing the residual error
\begin{equation}
\begin{split}
    \mathcal{L}_{PDE} = \mathbb{E}_{i,\mathbf{x},t,\mathbf{z}} \left[\left\vert\left\vert\frac{\partial G_\theta(\cdot)}{\partial t}- f\left(\mathbf{x},t,G_\theta(\cdot),\frac{\partial G_\theta(\cdot)}{\partial \mathbf{x}},\ldots,\frac{\partial^d G_\theta(\cdot)}{\partial \mathbf{x}^d}\right)\right\vert\right\vert_2^2\right]
\end{split}
\end{equation}
where $\cdot$ is a placeholder for $(\mathbf{x},t,i,\mathbf{z})$.

As noted before, the derivatives can readily be obtained via automatic differentiation.

Furthermore, we also force the true and approximated initial conditions to be equal by minimizing
\begin{equation}
    \mathcal{L}_{IC} = \mathbb{E}_{i,\mathbf{x},\mathbf{z}}\left[\left\vert\left\vert u(\mathbf{x},t=0;i) - G(\mathbf{x},t=0,i,\mathbf{z})\right\vert\right\vert_2^2\right].
\end{equation}
Note that $t$ is fixed here.

Next, we enforce the boundary conditions. This can vary from equation to equation. For Burger's equation we enforce the following
\begin{equation}
\begin{split}
    \mathcal{L}_{BC} =& \mathbb{E}_{t,i,\mathbf{z}}\left[\left\vert\left\vert G(\mathbf{x}_{UB},t,i,\mathbf{z})-G(\mathbf{x}_{LB},t,i,\mathbf{z})\right\vert\right\vert_2^2\right]\\&\qquad+\mathbb{E}_{t,i,\mathbf{z}}\left[\left\vert\left\vert\frac{\partial G(\mathbf{x}_{UB},t,i,\mathbf{z})}{\partial \mathbf{x}}-\frac{\partial G(\mathbf{x}_{LB},t,i,\mathbf{z})}{\partial \mathbf{x}}\right\vert\right\vert_2^2\right]
\end{split}
\end{equation}

where $\mathbf{x}_{UB}$ and $\mathbf{x}_{LB}$ denote the upper and lower boundary values of $\mathbf{x}$ respectively.

To train the reconstuctor $R$, we minimize
\begin{equation}
    \mathcal{L}_r = \mathbb{E}_{i,\mathbf{x},t,\mathbf{z}} \left[\left\vert\left\vert i-R(G(x,t,i,z))\right\vert\right\vert_2^2\right].
\end{equation}

Our overall objective function is therefore given by
\begin{equation}
    \mathcal{L} = \mathcal{L}_{GAN} + \alpha\mathcal{L}_{PDE} + \beta\mathcal{L}_{IC} + \gamma\mathcal{L}_{BC}+ \delta\mathcal{L}_r
\end{equation}
where $\alpha$, $\beta$ and $\gamma$ are scalars.

\begin{figure}[ht!]
  \centering
  \label{fig:results}
  \includegraphics[width=\textwidth]{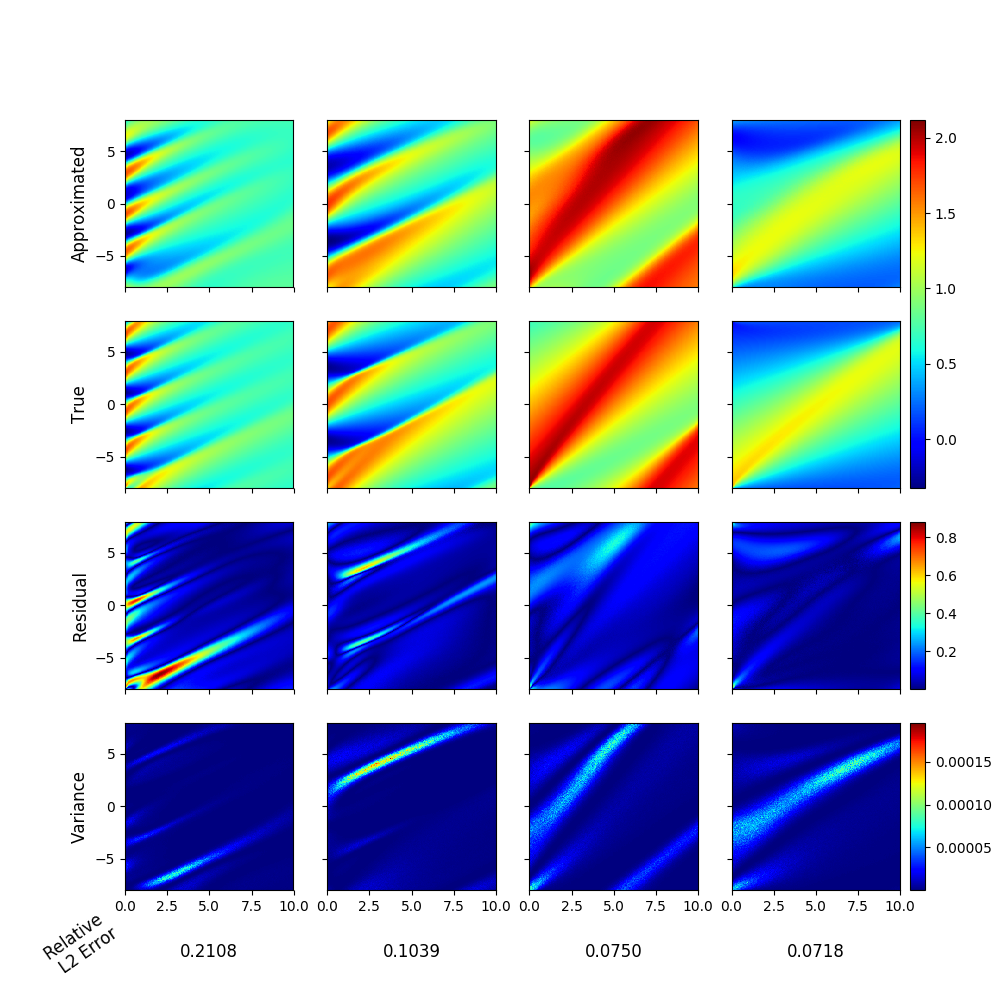}
  \caption{Results on some test fields.}
\end{figure}

\section{Experiments}
\label{exp}
In this section, we demonstrate our model on the Burger's equation which is given by
\begin{equation}
    \frac{\partial u}{\partial t} = 0.1\frac{\partial^2 u}{\partial x^2} - u\frac{\partial u}{\partial x}
\end{equation}
for initial conditions $u(x,t=0)$ of the form $a\sin(bx+c\pi)+d$ where $a$, $b$, $c$ and $d$ are real numbers.

We generate solutions for 120 different initial conditions by choosing different combinations of $a$, $b$, $c$ and $d$ from the sets $\{-1,1\}$, $\{0.1, 0.5, 0.9, 1.3, 1.7\}$, $\{0, 0.5, 0.9, 1.0\}$ and $\{0, 0.7, 1.4\}$ respectively. $20$ of these solutions constitute the test set (which is kept fixed for all experiments). The rest of the $100$ fields are randomly split for each experiment into training and validation sets in the ratio 85:15. The data is generated via the Chebfun package for MATLAB [\cite{chebfun}] for solving PDEs. We discretize $x \in \{-8,8\}$ into $512$ points and $t \in \{0,10\}$ such that the total number of points are around $100,000$.

We concatenate $u(x,t=0)$ at each of the $512$ values for $x$ into a vector. These are our initial conditions. We feed these initial conditions into our model as described in the previous section. Table \ref{tab:model-architecture} describes the model architecture. The reconstructor is fed the approximator's outputs at $64$ different spatio-temporal coordinates. We trained the model parameters using the Adam optimizer with a learning rate of $0.001$ for $1.6$ million iterations. At each iteration, we feed the model $64$ initial points to enforce $\mathcal{L}_{IC}$, $64$ boundary points to enforce $\mathcal{L}_{BC}$ and $512$ colocation points to enforce $\mathcal{L}_{PDE}$ and $\mathcal{L}_{GAN}$.

We achieved a relative $\ell_2$ error of $0.136$ on the test set. Figure \ref{fig:results} shows some of the fields generated by the model for initial conditions in the test set. Since the latent variable $z$ in the generator is a random variable, each time the model is run, it generates a slightly different field. As such, we also plot the variance in the model. Note that the variance loosely correlates with the residual (error) field and that both of these are high in regions close to discontinuities. This means that even though our model does not entirely approximate discontinuous regions correctly, it is nevertheless successful in identifying these areas.

While these results are preliminary, they do show the effectiveness of our method as a general PDE solver.
 
\begin{table}
  \caption{Model architecture}
  \label{tab:model-architecture}
  \centering
  \begin{tabular}{ccccc}
    \toprule
    & \multicolumn{3}{c}{Hidden}\\
    \cmidrule(r){2-4}
                    & Layers    & Neurons   &  Activation   & Output Neurons    \\
    \midrule
    Encoder         & $2$       & $128$     & ReLU          & $32$              \\
    Approximator    & $5$       & $256$     & Tanh          & $256$             \\
    Reconstructor   & $3$       & $256$     & Tanh          & $256$             \\
    Discriminator   & $3$       & $256$     & ReLU          & $256$             \\
    \bottomrule
  \end{tabular}
\end{table}

\section{Conclusion}
\label{conclusion}
We have demonstrated that a generative model can learn to predict solutions for unseen initial conditions when trained only a subset of them for a given PDE. This is the first step towards learning a general neural PDE solver which can provide reliable solutions for novel initial conditions. Furthermore, we have shown that as a result of the probabilistic nature of our model, uncertainty estimates also become freely available and, in our observation, correspond loosely with the error in the approximated solution of the PDE. 

\bibliography{iclr2020_conference}
\bibliographystyle{iclr2020_conference}

\end{document}